\documentclass[11pt]{article}
\usepackage[utf8]{inputenc}
\usepackage[T1]{fontenc}
\usepackage{amsmath,amssymb,amsfonts}
\usepackage{graphicx}
\usepackage{booktabs}
\usepackage{multirow}
\usepackage{hyperref}
\usepackage{url}
\usepackage{xcolor}
\usepackage{natbib}
\usepackage{tikz}
\usepackage{pgfplots}
\pgfplotsset{compat=1.18}
\usepackage[margin=1in]{geometry}
\usepackage{algorithm}
\usepackage{algorithmic}
\usepackage{float}
\usepackage{caption}
\usepackage{subcaption}

\title{GLiNER-Relex: A Unified Framework for Joint Named Entity Recognition and Relation Extraction}

\author{
Ihor Stepanov$^{1}$ \quad Oleksandr Lukashov$^{1}$ \quad Mykhailo Shtopko$^{1}$ \quad Vivek Kalyanarangan $^{2}$\\
$^{1}$Knowledgator Engineering, Kyiv, Ukraine \\
$^{2}$Baldor Technologies Pvt. Ltd. (IDfy), Mumbai, India \\
\texttt{ingvarstep@knowledgator.com}
}

\date{}

\begin{document}
\maketitle

\begin{abstract}
Joint named entity recognition (NER) and relation extraction (RE) is a fundamental task in natural language processing for constructing knowledge graphs from unstructured text. While recent approaches treat NER and RE as separate tasks requiring distinct models, we introduce GLiNER-Relex, a unified architecture that extends the GLiNER framework to perform both entity recognition and relation extraction in a single model. Our approach leverages a shared bidirectional transformer encoder to jointly represent text, entity type labels, and relation type labels, enabling zero-shot extraction of arbitrary entity and relation types specified at inference time. GLiNER-Relex constructs entity pair representations from recognized spans and scores them against relation type embeddings using a dedicated relation scoring module. We evaluate our model on four standard relation extraction benchmarks: CoNLL04, DocRED, FewRel, and CrossRE, and demonstrate competitive performance against both specialized relation extraction models and large language models, while maintaining the computational efficiency characteristic of the GLiNER family. The model is released as an open-source Python package with a simple inference API that allows users to specify arbitrary entity and relation type labels at inference time and obtain both entities and relation triplets in a single call. All models and code are publicly available.
\end{abstract}

\section{Introduction}

Information extraction (IE) from unstructured text is a foundational task in natural language processing (NLP), with broad applications in knowledge graph construction, question answering, document understanding, and retrieval-augmented generation (RAG) systems. Two of the most critical sub-tasks within IE are named entity recognition (NER)---the identification and classification of entities such as persons, organizations, and locations---and relation extraction (RE)---the detection and categorization of semantic relationships between identified entities.

Traditionally, NER and RE have been treated as separate tasks, addressed by pipeline approaches where entities are first extracted and then passed to a relation classifier \citep{zelenko2002kernel, roth2004linear}. While such pipelines are modular, they suffer from error propagation: mistakes in the NER stage cascade into the RE stage. This observation has motivated a substantial body of work on joint entity and relation extraction, where both tasks are modeled simultaneously to capture their interdependencies \citep{miwa2016end, zheng2017joint, fu2019graphrel}.

Recent advances in zero-shot NER, particularly the GLiNER framework \citep{zaratiana2024gliner}, have demonstrated that compact encoder-based models can achieve competitive performance on recognizing arbitrary entity types. GLiNER represents entity type labels and text tokens within a shared encoder, enabling flexible extraction of user-defined entity types at inference time. Subsequent work has extended this paradigm to multi-task information extraction \citep{stepanov2024gliner}, bi-encoder architectures for scalability \citep{stepanov2026million}, and biomedical adaptation \citep{yazdani2025gliner}.

However, the relation extraction component of the GLiNER ecosystem has received comparatively less attention. Existing approaches either rely on separate models, such as GLiREL \citep{boylan2025glirel}, which require pre-identified entities as input, or encode relations implicitly through label concatenation in the multi-task GLiNER formulation \citep{stepanov2024gliner}. Neither approach provides a truly unified model that jointly identifies entities and extracts relations in a single forward pass with shared representations.

In this paper, we introduce \textbf{GLiNER-Relex}, a unified architecture for joint NER and RE that extends the GLiNER framework with a dedicated relation extraction module. Our key contributions are:

\begin{itemize}
    \item \textbf{Unified architecture:} We propose a joint model that simultaneously performs entity recognition and relation extraction within a single encoder.
    \item \textbf{Zero-shot relation extraction:} GLiNER-Relex supports arbitrary entity and relation types specified through natural language labels.
    \item \textbf{Relation scoring mechanism:} We introduce a relation scoring module that was inspired by various knowledge graph embedding approaches.
    \item \textbf{Comprehensive evaluation:} We benchmark GLiNER-Relex on four standard RE datasets---CoNLL04, DocRED, FewRel, and CrossRE---comparing against GLiREL, GLiNER2, and GPT-5-mini.
    \item \textbf{Open-source release with simple API:} We release GLiNER-Relex as an open-source model with a straightforward Python API via the GLiNER package.
\end{itemize}

\section{Related Work}

\subsection{Named Entity Recognition}

Named entity recognition has evolved through several paradigms. Early rule-based systems \citep{appelt1993fastus} relied on hand-crafted patterns, while statistical methods such as conditional random fields (CRFs) \citep{lafferty2001conditional} introduced probabilistic sequence labeling. The advent of deep learning brought BiLSTM-CRF architectures \citep{lample2016neural}, which combined learned representations with structured prediction and became the dominant approach prior to the rise of pre-trained transformers. BERT-based models \citep{devlin2019bert} subsequently achieved state-of-the-art supervised NER by fine-tuning on task-specific labeled data.

However, all supervised NER models are limited to a fixed set of entity types defined during training. Zero-shot NER addresses this limitation by enabling the recognition of unseen entity types at inference time. InstructionNER \citep{wang2023instructionner} reformulated NER as a sequence generation task conditioned on natural language instructions, enabling LLMs to extract entities of specified types. UniversalNER \citep{zhou2024universalner} distilled ChatGPT annotations into a smaller model capable of recognizing diverse entity types across domains. GNER \citep{ding2024gner} further advanced generative NER by training on a large-scale dataset spanning diverse entity types.

\textbf{GLiNER} \citep{zaratiana2024gliner} took a different approach by defining NER as a matching problem between text spans and descriptions of the type of natural language entity within a shared bidirectional encoder, achieving competitive zero-shot performance at a fraction of the computational cost of LLM. Extensions of this framework include multi-task support for NER, RE, QA, and summarization \citep{stepanov2024gliner}; bi-encoder architectures for scaling to thousands of entity types \citep{stepanov2026million}; schema-driven extraction with GLiNER2 \citep{zaratiana2025gliner2}; synthetic data augmentation with NuNER \citep{bogdanov2024nuner}; and biomedical adaptation with GLiNER-BioMed \citep{yazdani2025gliner}.

\subsection{Relation Extraction}

Relation extraction approaches can be broadly categorized into pipeline, joint, and zero-shot methods.

\textbf{Pipeline approaches} first identify entities and then classify relations between entity pairs. Early work used kernel methods \citep{zelenko2002kernel} and feature engineering. PURE \citep{zhong2021pure} demonstrated that pipeline approaches with distinct contextual representations for entities and relations can achieve strong performance. However, pipeline methods suffer from error propagation between stages.

\textbf{Joint approaches} model entity recognition and relation extraction simultaneously using diverse paradigms. \textit{Sequence labeling methods} include Bi-LSTM with Tree-LSTM for relation prediction \citep{miwa2016end}, multi-tagging formulations \citep{zheng2017joint}, and position-attentive labeling for overlapping relations \citep{dai2019joint}. \textit{Decomposition-based methods} divide joint extraction into interdependent subtasks: CasRel \citep{wei2020casrel} maps head entities to tail entities via cascade binary tagging, \citet{yu2020joint} decompose extraction into head and tail entity stages, and PRGC \citep{zheng2021prgc} uses relation judgment, entity extraction, and subject--object alignment. \textit{Table-filling methods} such as UniRE \citep{wang2021unire} and TPLinker \citep{wang2020tplinker} treat extraction as filling word-pair tables with entity and relation labels. \textit{Set prediction methods} like SPN4RE \citep{sui2023spn4re} and OneRel \citep{shang2022onerel} formulate extraction as direct set prediction, avoiding sequential decoding errors. \textit{Span-based methods} like SpERT \citep{eberts2019spert} enumerate candidate spans and classify entity--relation combinations. \textit{Graph-based approaches} such as GraphRel \citep{fu2019graphrel} and GraphER \citep{zaratiana2024grapher} formulate IE as graph structure learning, while the autoregressive text-to-graph framework \citep{zaratiana2024autoregressive} takes a generative approach producing linearized graphs.

\subsection{Zero-Shot Relation Extraction}

Zero-shot relation extraction has attracted substantial attention as a means to overcome reliance on predefined relation taxonomies.

\textit{Entailment and reading comprehension approaches} reformulate RE as other well-studied tasks. \citet{levy2017zeroshot} reduced relation extraction to answering reading comprehension questions by associating natural-language questions with each relation slot. \citet{obamuyide2018zeroshot} and \citet{sainz2021label} reformulated relation extraction as a textual entailment task, using simple verbalizations of relation labels to leverage existing entailment models for zero-shot and few-shot settings.

\textit{Attribute and embedding learning approaches} project relations into semantic spaces. ZS-BERT \citep{chen2021zsbert} performs zero-shot relation classification by learning attribute representations for relation types, projecting both instances and unseen relation labels into a shared embedding space. ZSRE \citep{tran2023zsre} encodes text and relation labels separately, computing semantic correlations for each entity-label pair, achieving strong results but at limited efficiency. RE-Matching \citep{zhao2023rematching} proposes a fine-grained semantic matching method that decomposes relation representations into multiple components for more precise zero-shot matching.

\textit{Multiple-choice and template-based approaches} treat zero-shot RE as a classification problem. MC-BERT \citep{lan2023mcbert} models zero-shot relation classification as a multiple-choice problem, classifying entity pairs using previously unseen relation type labels. TMC-BERT \citep{moller2024tmcbert} extends this approach by incorporating entity type information and relation label descriptions for improved performance. However, both MC-BERT and TMC-BERT require constructing a separate input template for each entity pair and candidate label, which limits scalability.

\textit{Prompt-based and generative approaches} leverage language models for synthetic data and classification. RelationPrompt \citep{chia2022relationprompt} generates synthetic training examples at inference time using GPT-2, though it requires a large number of examples per label, making it resource-intensive. DSP \citep{lv2023dsp} employs discriminative soft prompts to jointly extract entities and relations in a zero-shot setting. ZS-SKA \citep{gong2024zska} performs zero-shot RE by using templates for data augmentation and incorporating an external knowledge graph.

\textit{LLM-based approaches} leverage large language models directly for relation extraction. \citet{li2024revisiting} demonstrated that meta in-context learning enables LLMs to achieve strong zero-shot and few-shot RE performance. For document-level RE, \citet{li2024lmrc} showed that combining a pre-trained classifier with LLaMA2 fine-tuned via LoRA yields significant improvements. GenRDK \citep{sun2024genrdk} uses chain-of-retrieval prompts with ChatGPT to generate synthetic data for fine-tuning.

\textit{Efficient encoder-based approaches} target both accuracy and scalability. GLiREL \citep{boylan2025glirel} adapted the GLiNER approach to relation classification, encoding relation labels alongside text in a shared bidirectional transformer and scoring entity-pair representations against relation-type embeddings. GLiREL achieved state-of-the-art results on Wiki-ZSL and FewRel while being significantly more efficient than template-based methods. However, GLiREL operates as a standalone relation classifier that requires pre-identified entities from an external NER model. GLiDRE \citep{armingaud2025glidre} extends the GLiNER approach to document-level relation extraction, achieving strong results on Re-DocRED. GLiNER2 treats relation extraction as a head-and-tail matching task after learning groups of relation representations. While it works without extracted entities, it can't be limited to selected entity types, making it an open-relation extraction approach.

\subsection{Joint Entity and Relation Extraction with Encoder Models}

The intersection of efficient encoder models and joint extraction remains underexplored. While GLiNER multi-task \citep{stepanov2024gliner} supports relation extraction by concatenating source entity and relation as a label (e.g., ``Bill Gates $\vert$ founded''), this formulation reduces RE to a span extraction problem and does not explicitly model entity pairs. GraphER \citep{zaratiana2024grapher} provides true joint extraction but operates in a supervised setting with fixed entity and relation types. Our work, GLiNER-Relex, bridges this gap by providing zero-shot joint NER and RE within a single efficient encoder model.

\section{Method}

\subsection{Overview}

GLiNER-Relex extends the GLiNER architecture to jointly perform named entity recognition and relation extraction. The model takes as input a text sequence along with user-specified entity type labels and relation type labels, and produces both entity spans with their types and relation triplets connecting entity pairs. The architecture consists of five main components: (1) a shared encoder that jointly processes text, entity labels, and relation labels; (2) a span representation layer for entity extraction; (3) an entity pair construction module with optional adjacency-guided pair selection; (4) a relation scoring layer; and (5) a multi-task training objective that jointly optimizes entity, adjacency, and relation losses.

\begin{figure}[H]
    \centering
    \includegraphics[width=0.8\linewidth]{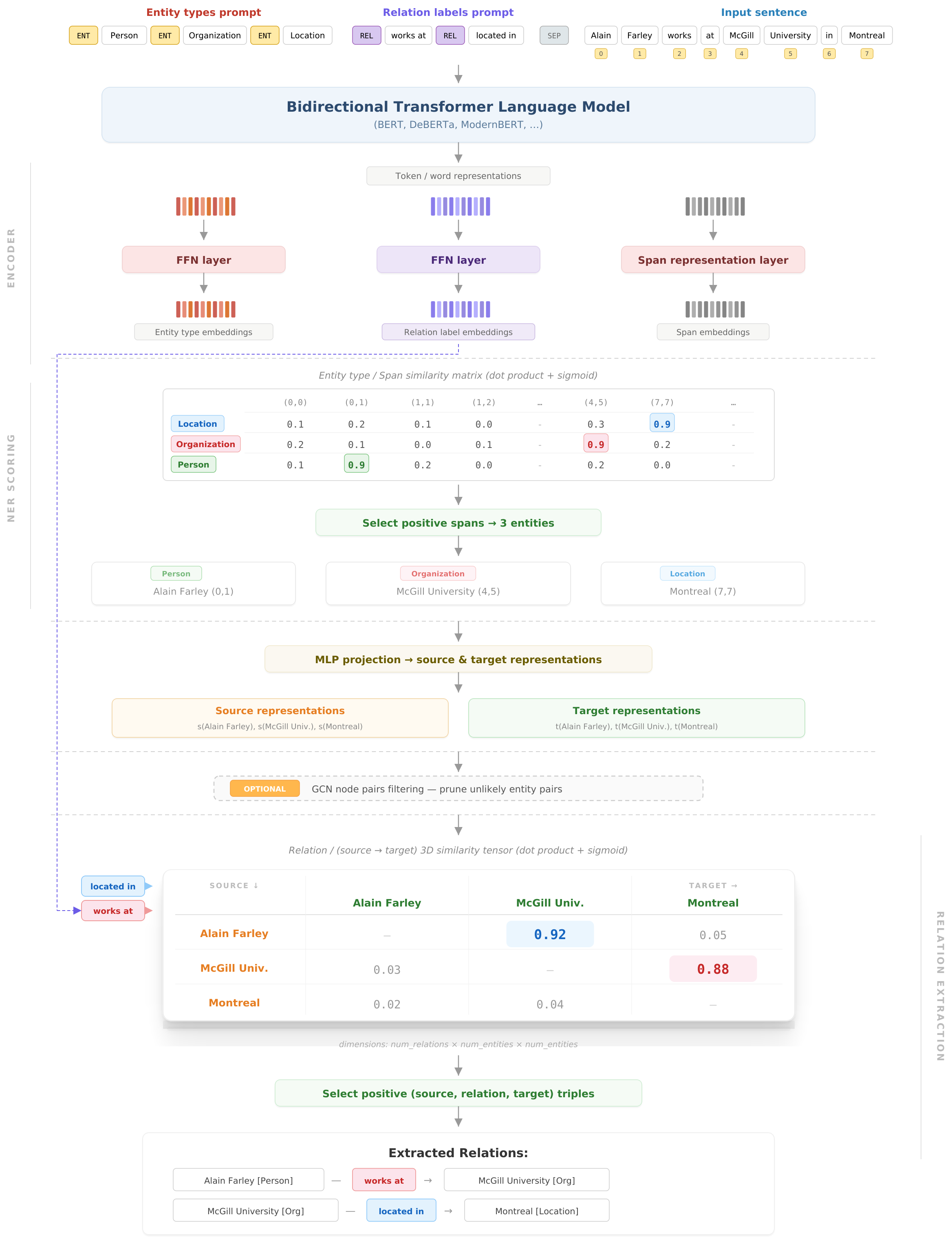}
    \caption{Overview of the GLiNER-Relex architecture.}
    \label{fig:gliner_relex}
\end{figure}

\subsection{Input Representation}

Given a text sequence $T = (t_0, t_1, \ldots, t_N)$, a set of entity type labels $\mathcal{Y}_E = \{e_1, e_2, \ldots, e_K\}$, and a set of relation type labels $\mathcal{Y}_R = \{r_1, r_2, \ldots, r_M\}$, we construct a unified input sequence by concatenating three prompted segments:

\begin{equation}
X = [\underbrace{\texttt{[ENT]}\; e_1 \;\texttt{[ENT]}\; e_2 \;\cdots\; \texttt{[ENT]}\; e_K}_{\text{entity types prompt}},\; \underbrace{\texttt{[REL]}\; r_1 \;\texttt{[REL]}\; r_2 \;\cdots\; \texttt{[REL]}\; r_M}_{\text{relation labels prompt}},\; \underbrace{\texttt{[SEP]}\; t_0\; t_1 \;\cdots\; t_N}_{\text{input sentence}}]
\end{equation}

\noindent Each entity type label is preceded by a special \texttt{[ENT]} delimiter token, and each relation type label is preceded by a special \texttt{[REL]} delimiter token. This layout places entity type labels and relation type labels into a shared context window with the input text, enabling cross-attention between all three components within the transformer encoder. The \texttt{[ENT]} and \texttt{[REL]} tokens' hidden representations after encoding serve as the entity type and relation type embeddings used for downstream scoring.

Throughout the paper, we use $\mathcal{Y}_E$ and $\mathcal{Y}_R$ to denote the sets of entity and relation \emph{type labels} provided at inference time, and $\mathcal{E}$ to denote the set of \emph{recognized entity spans} produced by the model (introduced in Section~\ref{sec:entity_pair}). These are distinct objects and are used consistently throughout.

\subsection{Shared Encoder}

The unified input sequence is processed by a bidirectional transformer encoder $f_{\text{enc}}$ (DeBERTa-v3 in our implementation):

\begin{equation}
H = f_{\text{enc}}(X)
\end{equation}

\noindent From the contextualized hidden states $H$, we extract three sets of representations:
\begin{itemize}
    \item \textbf{Word embeddings} $H_T = \{h_{t_i}\}_{i=0}^{N}$: Representations for each word in the input text, obtained by aggregating subword token embeddings.
    \item \textbf{Entity type embeddings} $H_E = \{h_{e_k}\}_{k=1}^{K}$: Representations for each entity type label, extracted at the positions of the corresponding \texttt{[ENT]} delimiter tokens.
    \item \textbf{Relation type embeddings} $H_R = \{h_{r_m}\}_{m=1}^{M}$: Representations for each relation type label, extracted at the positions of the corresponding \texttt{[REL]} delimiter tokens.
\end{itemize}

The word embeddings are optionally passed through a bidirectional LSTM layer for additional sequence modeling:

\begin{equation}
H_T' = \text{BiLSTM}(H_T)
\end{equation}

\subsection{Entity Extraction}

Following the standard GLiNER span-based approach, we construct span representations for all candidate spans up to a maximum width $W$:

\begin{equation}
s_{i,j} = \text{SpanRep}(H_T'[i:j])
\end{equation}

\noindent where $\text{SpanRep}$ combines start and end token representations with learned width embeddings. Entity type embeddings are projected through a dedicated layer:

\begin{equation}
\hat{h}_{e_k} = \text{Proj}_E(h_{e_k})
\end{equation}

Entity scores are computed via dot-product similarity between span and entity type representations:

\begin{equation}
\text{score}_{\text{ent}}(i, j, k) = s_{i,j} \cdot \hat{h}_{e_k}
\end{equation}

Entities are decoded using greedy span selection with a confidence threshold $\tau_E$.

\subsection{Entity Pair Construction}
\label{sec:entity_pair}

After entity extraction, the model must determine which pairs of recognized entities to evaluate for relations. Let $\mathcal{E}$ denote the set of recognized entities. GLiNER-Relex supports two entity pair construction strategies, selected via configuration.

\textbf{All-pairs enumeration.} The simplest approach enumerates all ordered pairs of recognized entities. Given $|\mathcal{E}|$ entities, this produces $|\mathcal{E}| \times (|\mathcal{E}| - 1)$ candidate pairs. While exhaustive, this strategy scales quadratically and is best suited for sentences with a moderate number of entities. This is the strategy used in the released GLiNER-Relex checkpoint (Section~\ref{sec:impl_details}).

\textbf{Adjacency-guided selection.} To reduce the number of candidate pairs, the framework optionally includes a \textit{RelationsRepLayer} that predicts a soft adjacency matrix $\hat{A} \in [0,1]^{|\mathcal{E}| \times |\mathcal{E}|}$ over entity span representations. A pair mask zeros out entries involving padded entities: $\hat{A}_{a,b} \leftarrow \hat{A}_{a,b} \cdot m_a \cdot m_b$. The layer supports six interchangeable decoder architectures:

\begin{itemize}
    \item \textit{Dot-product:} $\hat{A}_{a,b} = \sigma(s_a^\top s_b)$, optionally with $L_2$-normalization (cosine similarity). Parameter-free baseline.
    \item \textit{Bilinear:} Projects entities via $W_P \in \mathbb{R}^{D \times d_L}$ and scores $\hat{A}_{a,b} = \sigma(z_a^\top z_b)$ where $z = W_P^\top s$, decoupling adjacency from span representations.
    \item \textit{MLP:} Concatenates pairs and applies a two-layer MLP, $\hat{A}_{a,b} = \sigma(\text{MLP}([s_a;s_b]))$, enabling asymmetric and nonlinear interactions.
    \item \textit{Attention:} Multi-head self-attention over entities, with attention weights averaged across heads to form $\hat{A}$.
    \item \textit{GCN:} Computes an initial dot-product adjacency, applies a graph convolutional layer with symmetric normalization ($\tilde{D}^{-1/2}\tilde{A}\tilde{D}^{-1/2}$) to refine representations via message-passing, then predicts the final adjacency from the updated features.
    \item \textit{GAT:} Multi-head attention updates entity representations, which are then projected and scored bilinearly, combining contextual refinement with a learnable output space.
\end{itemize}

Entity pairs with $\hat{A}_{a,b}$ above a threshold $\tau_A$ are retained for relation classification, effectively pruning unlikely pairs before the more expensive relation scoring step. During training with ground-truth adjacency labels, this component is supervised with a dedicated adjacency loss (Section~\ref{sec:training_objective}).

The six decoders described above are framework-level options supported by the GLiNER-Relex codebase. The released checkpoint uses the all-pairs enumeration strategy and does not activate any adjacency decoder; a systematic ablation of decoder architectures is left to future work (Section~\ref{sec:limitations}).

For each selected entity pair $(a, b)$, the head and tail span representations $s_a$ and $s_b$ are extracted from the entity representations for downstream relation scoring.

\subsection{Relation Scoring}
\label{sec:relation_scoring}

Given selected entity pairs with head representations $s_a$ and tail representations $s_b$, along with relation type embeddings $H_R = \{h_{r_m}\}_{m=1}^{M}$ from the shared encoder, the model scores each entity pair against each candidate relation type. The GLiNER-Relex framework implements two families of relation scoring mechanisms.

\textbf{Pair representation layer.} In this approach, head and tail entity representations are concatenated and projected through an MLP layer to produce a unified pair representation:

\begin{equation}
p_{a,b} = \text{MLP}\!\left([s_a ; s_b]\right)
\end{equation}

\noindent where $[s_a ; s_b] \in \mathbb{R}^{2D}$ is the concatenation and $\text{MLP}: \mathbb{R}^{2D} \to \mathbb{R}^{D}$ projects the pair back to the shared embedding dimension via a linear layer with dropout. The relation score is then computed as a dot product between the pair representation and the relation type embedding:

\begin{equation}
\text{score}_{\text{rel}}(a, b, m) = p_{a,b} \cdot h_{r_m}
\end{equation}

This formulation encourages the model to learn a shared semantic space in which entity pair representations are close to the embeddings of their corresponding relation types. Since both entity pairs and relation labels are encoded jointly by the shared transformer, the dot-product scoring enables zero-shot generalization to unseen relation types specified through natural language descriptions at inference time.

\textbf{Knowledge graph--inspired triple scoring layers.} As an alternative, the framework supports a family of triple scoring functions $f(s_a, h_{r_m}, s_b) \to \mathbb{R}$ drawn from the knowledge graph embedding literature \citep{bordes2013transe, yang2015distmult, trouillon2016complex}. Each scoring function models the interaction between head entity, relation, and tail entity representations using a distinct geometric or algebraic assumption. The implemented variants include:

All triple scoring variants operate on the same entity and relation representations produced by the shared encoder. Each scoring function receives the head, relation, and tail embeddings and produces a scalar compatibility score, computed over all entity pair--relation type combinations in a single batched operation.

In our experiments, we found that the pair representation layer with MLP projection achieves the best balance of accuracy and efficiency, and it is used in the released model. The knowledge graph--inspired layers offer a richer space of inductive biases that may benefit specialized applications, such as settings where relation symmetry, transitivity, or compositionality are important structural priors.

\subsection{Training Objective}
\label{sec:training_objective}

The model is trained with a multi-task objective that combines up to three loss components:

\begin{equation}
\mathcal{L} = \lambda_E \mathcal{L}_{\text{ent}} + \lambda_A \mathcal{L}_{\text{adj}} + \lambda_R \mathcal{L}_{\text{rel}}
\end{equation}

\noindent where $\mathcal{L}_{\text{ent}}$ is the entity extraction loss, $\mathcal{L}_{\text{adj}}$ is the optional adjacency matrix loss (present only when the adjacency-guided pair selection is used), and $\mathcal{L}_{\text{rel}}$ is the relation classification loss. The coefficients $\lambda_E$, $\lambda_A$, and $\lambda_R$ control the relative contribution of each component. The specific values used for the released checkpoint are reported in Section~\ref{sec:impl_details}.

All losses use focal loss \citep{lin2017focal} with optional negative sampling to handle the severe class imbalance inherent in both tasks:

\begin{equation}
\mathcal{L}_{\text{focal}}(p, y) = -\alpha (1-p_t)^{\gamma} \log(p_t)
\end{equation}

\noindent where $p_t = p$ if $y=1$ and $p_t = 1-p$ otherwise, $\alpha$ is a balancing factor, and $\gamma$ is the focusing parameter. When $\gamma = 0$ the focal loss reduces to $\alpha$-balanced binary cross-entropy; we retain the focal-loss interface so that $\gamma$ can be increased in future training runs without changes to the optimization pipeline.

\subsection{Training Data}

Following the synthetic data generation paradigm established in prior GLiNER work \citep{stepanov2024gliner}, we construct training data through a two-stage annotation pipeline using LLMs applied to text sampled from FineWeb \citep{penedo2024fineweb}.

\textbf{Stage 1: Large-scale pre-training data.} We sampled a large corpus of text from FineWeb, segmented it into individual sentences, and annotated approximately 1 million sentences using Qwen3-32B \citep{yang2025qwen3}. In addition, we annotated 50,000 full-length texts (without sentence splitting) to expose the model to document-level context. The sentence-level and document-level annotations were then mixed into a unified training set. Each training instance contains tokenized text, entity spans with type labels, and relation triplets linking entity pairs with relation type labels, aligned with the GLiNER input format.

\textbf{Stage 2: High-quality fine-tuning data.} To further improve model quality, we curated a smaller, high-quality dataset of approximately 3,000 examples using a multi-step annotation pipeline. First, we sampled texts from FineWeb and applied a general-purpose NER model to extract coarse entity types (e.g., person, organization, location). We filtered for entity-rich passages, retaining only texts with a sufficient density of recognized entities. Next, the selected texts were annotated using Gemini \citep{google2025gemini} in two passes: (1)~an extraction pass, where the model was prompted to identify entities and relations, with relations drawn from 12 pre-defined semantic groups (e.g., affiliation, spatial, causal) that serve as high-level categories without strictly limiting the verbalized relation labels; and (2)~a correction pass, where the model reviewed and refined the extracted entities and relations to improve annotation consistency and accuracy. This high-quality dataset was used for final fine-tuning of the model.

\subsection{Implementation and Training Details}
\label{sec:impl_details}

The released GLiNER-Relex checkpoint uses a DeBERTa-v3-large shared encoder followed by a bidirectional LSTM with hidden size 1024. The model supports input sequences of up to 2048 words and candidate spans of up to $W = 12$ words, which covers the vast majority of entity mentions in the training data.

For entity-pair construction (Section~\ref{sec:entity_pair}), the released model uses the \emph{all-pairs enumeration} strategy; the adjacency-guided selection path and the six decoder architectures described in Section~\ref{sec:entity_pair} are supported by the framework but are not active in the released checkpoint. Accordingly, the adjacency loss is omitted from the training objective ($\lambda_A = 0$) and the effective objective reduces to $\mathcal{L} = \lambda_E\,\mathcal{L}_{\text{ent}} + \lambda_R\,\mathcal{L}_{\text{rel}}$ with $\lambda_E = \lambda_R = 1.0$. Both component losses use the focal-loss formulation of Section~\ref{sec:training_objective} with $\alpha = 0.75$ and $\gamma = 0$, which reduces to $\alpha$-balanced binary cross-entropy.

Training follows the two-stage pipeline described in Section~3.8. \textbf{Stage~1} (large-scale pre-training on the approximately one million synthetically annotated sentences plus 50{,}000 document-level examples) runs for a single epoch with AdamW, a warmup ratio of 0.05, batch size 8, and differential learning rates of $1\!\times\!10^{-5}$ for the encoder and $5\!\times\!10^{-5}$ for the task-specific layers (span representation, pair representation, and relation projection). \textbf{Stage~2} (fine-tuning on the approximately 3{,}000 high-quality Gemini-annotated examples) runs for 5 epochs with the same optimizer, warmup ratio, and batch size, but with reduced learning rates of $3\!\times\!10^{-6}$ for the encoder and $5\!\times\!10^{-6}$ for the task-specific layers. The lower Stage-2 learning rates are chosen to preserve the broad zero-shot capabilities learned in Stage~1 while refining the model on the higher-quality annotations.

At inference, we use an entity confidence threshold $\tau_E = 0.3$ and a relation confidence threshold $\tau_R = 0.5$, consistent with the default values exposed in the public API (Section~\ref{sec:usage}). Table~\ref{tab:hyperparameters} summarizes all hyperparameters for reproducibility.

\begin{table}[H]
\centering
\small
\begin{tabular}{llr}
\toprule
\textbf{Component} & \textbf{Hyperparameter} & \textbf{Value} \\
\midrule
\multirow{3}{*}{Encoder}
  & Backbone                    & DeBERTa-v3-large \\
  & Max sequence length         & 2048 words \\
  & BiLSTM hidden size          & 1024 \\
\midrule
Span encoder
  & Max span width $W$          & 12 words \\
\midrule
\multirow{2}{*}{Pair construction}
  & Strategy                    & all-pairs enumeration \\
  & Adjacency decoder           & none \\
\midrule
\multirow{5}{*}{Loss}
  & Focal $\alpha$              & 0.75 \\
  & Focal $\gamma$              & 0 \\
  & $\lambda_E$ (entity)        & 1.0 \\
  & $\lambda_A$ (adjacency)     & 0 (disabled) \\
  & $\lambda_R$ (relation)      & 1.0 \\
\midrule
\multirow{6}{*}{Stage~1 (pre-training)}
  & Optimizer                   & AdamW \\
  & Encoder learning rate       & $1\!\times\!10^{-5}$ \\
  & Task-specific layers learning rate    & $5\!\times\!10^{-5}$ \\
  & Warmup ratio                & 0.05 \\
  & Batch size                  & 8 \\
  & Epochs                      & 1 \\
\midrule
\multirow{6}{*}{Stage~2 (fine-tuning)}
  & Optimizer                   & AdamW \\
  & Encoder learning rate       & $3\!\times\!10^{-6}$ \\
  & Task layer learning rate    & $5\!\times\!10^{-6}$ \\
  & Warmup ratio                & 0.05 \\
  & Batch size                  & 8 \\
  & Epochs                      & 5 \\
\midrule
\multirow{2}{*}{Inference}
  & Entity threshold $\tau_E$   & 0.3 \\
  & Relation threshold $\tau_R$ & 0.5 \\
\bottomrule
\end{tabular}
\caption{Hyperparameters for the released GLiNER-Relex checkpoint.}
\label{tab:hyperparameters}
\end{table}

\section{Experiments}

\subsection{Benchmarks}

We evaluate GLiNER-Relex on four standard relation extraction benchmarks:

\textbf{CoNLL04} \citep{carreras2004conll04}: A dataset from news articles annotated with entity types (Person, Organization, Location, Other) and relation types (Located-In, Work-For, OrgBased-In, Live-In, Kill). It is commonly used for evaluating joint entity and relation extraction models.

\textbf{DocRED} \citep{yao2019docred}: A large-scale document-level relation extraction dataset constructed from Wikipedia. It contains 96 relation types and requires reasoning across multiple sentences within a document to identify relations, making it substantially more challenging than sentence-level benchmarks.

\textbf{FewRel} \citep{han2018fewrel}: A few-shot relation classification benchmark with 100 relation types derived from Wikidata. We evaluate in the zero-shot setting where the model must classify relations for types not seen during training.

\textbf{CrossRE} \citep{bassignana2022crossre}: A cross-domain relation extraction dataset spanning multiple domains (AI, literature, music, politics, science, news). It is designed to evaluate the transferability of RE models across different textual domains.

\subsection{Baselines}

We compare GLiNER-Relex against:

\begin{itemize}
    \item \textbf{GLiREL} \citep{boylan2025glirel}: A GLiNER-family model specialized for zero-shot relation classification. GLiREL operates on pre-identified entities rather than performing end-to-end extraction; we evaluate it by supplying ground-truth entity spans and types from each benchmark, so the model predicts only which relation (if any) holds over each entity pair. This gives GLiREL an upper-bound-style advantage relative to joint systems, and we include it to characterize the performance ceiling of specialized relation classifiers when NER errors are fully removed.
    \item \textbf{GLiNER2} \citep{zaratiana2025gliner2}: The multi-task GLiNER model from Fastino Labs, which supports NER, text classification, and structured extraction. We evaluate its relation extraction capability through the schema-driven interface.
    \item \textbf{GPT-5-mini}: OpenAI's compact large language model, evaluated in a zero-shot prompting setup with structured output specifications for relation extraction.
\end{itemize}

\subsection{Evaluation Protocol}

All models are evaluated in a zero-shot setting where no training examples from the target datasets are provided. We report Micro-F1 scores for relation extraction, measuring the overall precision-recall balance across all relation types. For GLiNER-Relex, inputs exceeding the model's maximum sequence length of 2048 words (relevant primarily for a small number of long DocRED documents) are truncated at the document end; no sliding-window aggregation is performed, so relations spanning beyond the truncation point are not recoverable.

Entity handling differs across systems and is made explicit in Table~\ref{tab:results}. GLiNER-Relex and GLiNER2 perform end-to-end extraction from raw text. GPT-5-mini receives the list of candidate entity types in the prompt and produces entities and relations jointly in its structured output. GLiREL, as a dedicated relation classifier, is evaluated with ground-truth entity spans and types supplied as input; its scores therefore reflect relation-classification performance conditional on perfect NER rather than end-to-end extraction.

\subsection{Results}

Table~\ref{tab:results} presents the zero-shot relation extraction results across all four benchmarks.

\begin{table}[H]
\centering
\begin{tabular}{llccccc}
\toprule
\textbf{Model} & \textbf{Entities} & \textbf{CoNLL04} & \textbf{DocRED} & \textbf{FewRel} & \textbf{CrossRE} & \textbf{Avg.} \\
\midrule
GLiREL$^{\dagger}$ & gold & 4.5 & 2.4 & \textbf{24.0} & 1.4 & 8.1 \\
GLiNER2 & predicted & 32.9 & 11.7 & 20.8 & 6.0 & 17.8 \\
GPT-5-mini & prompted & \textbf{42.4} & 18.6 & 15.0 & 12.4 & 22.1 \\
\midrule
GLiNER-Relex (ours) & predicted & 40.4 & \textbf{31.3} & 12.5 & \textbf{18.1} & \textbf{25.6} \\
\bottomrule
\end{tabular}
\caption{Zero-shot relation extraction performance (Micro-F1 \%) on four benchmarks. Bold indicates the best performance per dataset across end-to-end systems. The ``Entities'' column indicates how each model obtains entity spans at inference time: \emph{predicted} (joint extraction from raw text), \emph{prompted} (LLM extracts entities and relations jointly from natural-language prompt), or \emph{gold} (ground-truth entities supplied as input). The ``Avg.'' column reports the unweighted mean Micro-F1 across the four benchmarks. $^{\dagger}$GLiREL is a dedicated relation classifier evaluated with ground-truth entity spans and types, and is not directly comparable to the end-to-end systems; we include it as an upper-bound reference for specialized relation classification.}
\label{tab:results}
\end{table}

GLiNER-Relex demonstrates strong performance across the evaluated benchmarks, achieving the best Micro-F1 on two out of four datasets among end-to-end systems and competitive results on the remaining two.

On \textbf{CoNLL04}, GLiNER-Relex achieves 40.4\% Micro-F1, closely approaching GPT-5-mini (42.4\%) and substantially outperforming GLiNER2 (34.1\%). This demonstrates that our unified architecture can effectively capture sentence-level relations while maintaining the efficiency advantages of encoder-based models.

On \textbf{DocRED}, GLiNER-Relex achieves the highest performance at 31.3\%, outperforming GPT-5-mini (18.6\%) by a large margin of 12.7 percentage points and GLiNER2 (12.4\%) by 18.9 points. This result is particularly significant because DocRED requires document-level reasoning across multiple sentences, suggesting that the shared encoder representations in GLiNER-Relex effectively capture long-range dependencies between entities.

On \textbf{FewRel}, GLiNER-Relex achieves 12.5\%, trailing both GPT-5-mini (15.0\%) and GLiNER2 (16.8\%). GLiREL, which is evaluated with gold entities and was designed specifically for the sentence-level, entity-pair-aligned format that FewRel exemplifies, achieves the highest score at 23.9\%. The relatively lower performance of end-to-end systems here reflects two factors: FewRel's 100 fine-grained Wikidata relation types challenge the zero-shot generalization of the relation-type embeddings, and the benchmark's design favors models with access to pre-identified entity pairs---an advantage that GLiREL receives by construction.

On \textbf{CrossRE}, GLiNER-Relex achieves the best performance at 18.1\%, substantially outperforming GPT-5-mini (12.4\%), GLiNER2 (4.9\%), and GLiREL (1.4\%). This result highlights the model's strong cross-domain transferability, a direct benefit of the zero-shot formulation where relation types are specified through natural language descriptions.

A consistent pattern emerges from the GLiREL results: despite the advantage of receiving ground-truth entities, GLiREL's performance drops sharply outside of FewRel's narrow sentence-level format---to 4.5\% on CoNLL04, 2.2\% on DocRED, and 1.4\% on CrossRE. This suggests that the performance gap between specialized and unified relation extractors is not primarily driven by NER errors but by the distributional and structural assumptions baked into each architecture. Joint modeling with broad zero-shot pre-training, as in GLiNER-Relex, transfers more reliably across domains and granularities.

Restricting the comparison to end-to-end systems for apples-to-apples averaging, GLiNER-Relex's mean Micro-F1 across the four benchmarks is 25.6\%, compared to 22.1\% for GPT-5-mini and 17.1\% for GLiNER2. The model particularly excels on benchmarks requiring document-level reasoning (DocRED) and cross-domain generalization (CrossRE).

\section{Discussion}

\subsection{Key Benefits}

GLiNER-Relex unifies capabilities that prior approaches in the GLiNER ecosystem offer only in isolation. Where GraphER \citep{zaratiana2024grapher} requires supervised training on fixed types, GLiREL \citep{boylan2025glirel} depends on an external NER pipeline, and GLiNER multi-task \citep{stepanov2024gliner} reduces RE to span extraction via label concatenation, GLiNER-Relex jointly extracts entities and relations in a single forward pass with explicit entity-pair modeling and zero-shot generalization to arbitrary types.

This design yields three practical benefits. First, the unified architecture eliminates error propagation between separate NER and RE stages, as both tasks share representations within the same encoder. Second, the zero-shot formulation allows users to adapt the model to new domains by simply specifying entity and relation type labels as natural language strings, without any retraining. Third, the encoder-based architecture is orders of magnitude faster than autoregressive LLM-based extraction, making it well suited for latency-sensitive and resource-constrained deployments. The benchmark results confirm these advantages: GLiNER-Relex achieves the highest average Micro-F1 across all four datasets (25.6\% vs.\ 22.1\% for GPT-5-mini and 17.1\% for GLiNER2), with particularly strong performance on document-level (DocRED) and cross-domain (CrossRE) tasks.

To quantify the efficiency advantage of the encoder-based design, we benchmarked GLiNER-Relex against GPT-5-mini on a held-out set of 50 documents sampled from FineWeb (mean length 288 words, approximately 360 sub-word tokens), annotated with a DocRED-style schema comprising 6 entity types and 50 relation labels. GLiNER-Relex was executed on a single NVIDIA L4 GPU with batch size one; GPT-5-mini was accessed through the OpenAI API with default reasoning effort and structured JSON output, and per-request timings include network latency. Under these conditions, GLiNER-Relex achieved a mean per-document latency of 0.9~s (throughput 1.11 docs/sec), while GPT-5-mini averaged 64~s per document (throughput 0.016 docs/sec)---a throughput advantage of approximately $70\times$ in favor of the encoder-based model. This gap reflects both architectural differences and the reasoning-token overhead inherent to GPT-5-mini; reducing the reasoning effort would narrow the margin at some cost to extraction quality. The same workload incurs a non-trivial per-token cost on the hosted API, while GLiNER-Relex runs entirely on local infrastructure. These measurements substantiate the efficiency claim quantitatively and make GLiNER-Relex well-suited to high-volume pipelines such as knowledge-graph construction over corpora, where per-document extraction cost compounds rapidly across millions of documents.

\subsection{Usage}
\label{sec:usage}

GLiNER-Relex is released as an open-source model with a simple Python API that enables joint entity and relation extraction with user-defined types:

\begin{verbatim}
from gliner import GLiNER

model = GLiNER.from_pretrained(
    "knowledgator/gliner-relex-large-v1.0"
)

entity_labels = ["location", "person", "date", "structure"]
relation_labels = ["located in", "designed by", "completed in"]

text = ("The Eiffel Tower, located in Paris, France, "
        "was designed by engineer Gustave Eiffel "
        "and completed in 1889.")

entities, relations = model.inference(
    texts=[text],
    labels=entity_labels,
    relations=relation_labels,
    threshold=0.3,
    relation_threshold=0.5,
    return_relations=True,
    flat_ner=False
)
\end{verbatim}

The \texttt{inference} method accepts entity and relation type labels as plain strings, with separate confidence thresholds for entity recognition (\texttt{threshold}) and relation extraction (\texttt{relation\_threshold}). Setting \texttt{return\_relations=True} enables joint extraction, returning both entity spans with types and relation triplets linking entity pairs. The \texttt{flat\_ner} parameter controls whether overlapping entity spans are permitted.

\subsection{Applications to GraphRAG}

A particularly promising application is in Graph-based Retrieval-Augmented Generation (GraphRAG) pipelines \citep{edge2025localglobalgraphrag}. GraphRAG constructs a knowledge graph from a document corpus and leverages graph structure---community detection, summarization, multi-hop traversal---to answer complex queries that require synthesizing information across passages. Current implementations typically rely on LLMs for extraction, which introduces significant computational cost at scale.

GLiNER-Relex offers an attractive alternative in this setting: as an encoder-based model, it is substantially faster than autoregressive LLMs (quantified in Section~5.1), which could enable knowledge graph construction over large corpora within tighter time and cost budgets. Its zero-shot capabilities allow domain-specific extraction without fine-tuning. We view GLiNER-Relex as a promising extraction backbone for GraphRAG systems in latency-sensitive, resource-constrained, or privacy-sensitive deployment scenarios, and the same model can be reused at query time to parse user questions into entity mentions for graph traversal. A systematic comparison of GraphRAG pipelines built on GLiNER-Relex versus LLM-based extractors---measuring end-to-end answer quality, graph coverage, and total cost---is an interesting direction that we leave to future work.

\subsection{Limitations and Future Directions}
\label{sec:limitations}

GLiNER-Relex has several limitations. The zero-shot performance does not yet match fully supervised models or frontier LLMs on all benchmarks. Our experiments indicate room for improvement, particularly with large numbers of fine-grained relation types where the embedding space may require hierarchical or prototype-based representations.

Precision degrades in entity-dense passages: all-pairs enumeration produces quadratic candidate pairs, increasing spurious predictions. Although adjacency-guided selection mitigates this, dense entity graphs remain challenging for domains such as biomedicine, legal text, and financial reports.

Long document extraction is limited by two compounding factors. First, many recent encoder models are pre-trained on sequences shorter than 512 tokens, which limits their generalization to longer inputs. Second, as document length grows, so does the number of recognized entities, and the number of candidate entity pairs increases quadratically. This places increasing pressure on the model's fixed-dimensional embedding space, which must encode rich semantic and relational information for a rapidly growing set of combinations.

These limitations suggest several future directions: (1) extending the bi-encoder architecture \citep{stepanov2026million} to decouple the type vocabulary from the input context; (2) hierarchical encoding or cross-chunk attention for long documents; (3) integration with entity linking through GLinker for end-to-end knowledge graph construction; (4) a systematic ablation of the six adjacency-decoder architectures described in Section~\ref{sec:entity_pair}, which are supported by the framework but inactive in the current released checkpoint; and (5) end-to-end evaluation of GraphRAG pipelines built on GLiNER-Relex versus LLM-based extractors.

\section{Conclusion}

We introduced GLiNER-Relex, a unified framework for joint named entity recognition and relation extraction that extends the GLiNER architecture with a dedicated relation extraction module. Our model achieves competitive zero-shot performance on standard relation extraction benchmarks, outperforming both GLiNER2 and GPT-5-mini on document-level and cross-domain benchmarks, while maintaining the computational efficiency characteristic of encoder-based models. The model's simple inference API allows users to specify arbitrary entity and relation types through natural language labels, enabling zero-shot extraction across diverse domains without retraining. GLiNER-Relex provides an efficient and accessible solution for extracting structured knowledge from unstructured text, with applications spanning knowledge graph construction, document understanding, and information extraction across domains including biomedicine, law, enterprise knowledge management, and financial intelligence. Being much more efficient than LLMs and having competitive zero-shot capabilities make GLiNER-Relex a promising choice for building knowledge graphs that power RAG systems, enabling multi-hop question answering.

\bibliographystyle{plainnat}
\bibliography{main}

\end{document}